# AUTOMATING LEUKEMIA DIAGNOSIS WITH AUTOENCODERS: A COMPARATIVE STUDY

Minoo Sayyadpour*

*Department of Math and Computer Scienc, Kharazmi University*

*Tehran, Iran*

*std_minoosayyadpour@khu.ac.ir*

Nasibe Moghaddamniya

*Department of Computer Engineering, Science and Research Branch, Islamic Azad University*

*Tehran, Iran*

*nasibe.moghaddamniya@gmail.com*

Touraj Banirostam

*Department of Computer Engineering, Central Tehran Branch, Islamic Azad University*

*Tehran, Iran*

*banirostam@iauctb.ac.ir*

Leukemia is one of the most common and death-threatening types of cancer that threaten human life. Medical data from some of the patient's critical parameters contain valuable information hidden among these data. On this subject, deep learning can be used to extract this information. In this paper, AutoEncoders have been used to develop valuable features to help the precision of leukemia diagnosis. It has been attempted to get the best activation function and optimizer to use in AutoEncoder and designed the best architecture for this neural network. The proposed architecture is compared with this area's classical machine learning models. Our proposed method performs better than other machine learning in precision and f1-score metrics by more than 11%.

## 1. Introduction

Cancer is a pervasive ailment that transcends temporal and geographical boundaries. It stands as the second leading cause of mortality on a global scale and the third leading cause of death specifically within the nation of Iran. Projections indicate an anticipated escalation in the worldwide incidence rate of this affliction in the years ahead. Notably, leukemia occupies the fifth position in terms of prevalence worldwide and holds the second position within Iran [1]. Accurate and on-time recognition of this disease helps patients to be able to perform better treatments at earlier times and prevent the spread of cancerous cells and their spread to other organs [2]. Also, some types of cancer, such as leukemia, can be treated if detected on time. In the same way, increasing the accuracy of diagnosis can help save many people's lives. The process of recognizing cancer in the screening stage with methods such as CXR, mammography, sonography, endoscopy, and CT scan, compared to simpler blood and urine test methods, is expensive. Among the unknown aspects of cancer diagnosis, we can mention the following:

- How many samples are needed to train the detection system to achieve the desired accuracy, and to what extent can this data be accessed?
- Can other data be used with this technique to diagnose other diseases and other types of cancer?
- What are the parameters in the problem that are more important?
- To what extent can the doctor and the patient rely on these results?

In each sample (human), many parameters can probably be classified after some quantitative measurement to determine whether a person has cancer. However, finding compelling features in cancer diagnosis is a challenging process. Sometimes it is impossible to access an expert in the desired field (doctor, physician, etc.).

Deep learning automatically performs the feature extraction and learns which features are effective. This process is called Feature Extraction [2]. Deep learning can also perform classification based on features, and achieve higher accuracy in classification. These benefits have led to using deep learning in cancer diagnosis. Meanwhile, AutoEncoders can reduce features and extract efficient and valuable features; abstractly, they still maintain the main frame of the input sample. This led us to use Autoencoders to diagnose cancer. This paper has contributions that can be stated in the following:

- Using tabular data in AutoEncoders for cancer recognition.
- Use of Leukemia dataset of Sina Hospital in Hamadan.
- Find the best optimization algorithm and activation function for AutoEncoder networks.

The structure of remaining sections of this paper is organized as follows:
The second section reviews some essential methods in machine learning and deep learning networks. The third section deals with the research methodology. This section presents a new Leukemia detection method using AutoEncoder, showing the best results in accuracy and precision among the previous methods. The fourth chapter evaluates the results and compares them with each other. The fifth chapter concludes the study results.

## 2. Backgrounds

Neural networks can extract patterns and identify features difficult for humans to identify with their remarkable ability to derive results from complex data. This section discusses the studies conducted on the studies conducted in this field.

Medical Decision Support System (MDSS) helps doctors make decisions and accurately diagnose diseases. In 1980, MDSS was mentioned and brought about a revolution in decision support systems [3 and 4]. These systems help doctors to diagnose and track diseases and reduce possible errors. In [4], researchers designed a medical assistant system to help radiologists identify hylic tumors based on a distributed architecture with three specific nodes: Radiologist Visual Interface, Support services information system, and Web-based decision. Another medical assistant system to help asthma patients was presented by [5], which helps doctors control this chronic disease. In [6], they designed an MDSS to identify types of hearing problems, which helps experts control and solve them. In [7], another medical decision support system has been implemented to control Leukemia. In this system, in order to diagnose and predict Leukemia, four different classification methods have been used:

- Rule Based Reasoning
- Case-Based Reasoning
- Neural Network
- Discriminant Analysis

Many investigations have been done in diagnosing and classifying all types of cancer. Many of them use machine learning algorithms to classify data. In these references, in order to reduce the size of training data and define appropriate subsets of input variables, the feature extraction approach has been used. Feature extraction is the selection of a subset of input variables that increases the classifier's efficiency. In [8], they used a better diagnostic tool called Fine Needle Aspiration Cytology (FNAC) to classify the pattern of breast cancer. In this regard, Multivariate Adaptive Regression Splines (MARS) detect the appropriate subset of input variables from a neural network model. Therefore, the classification accuracy is increased in the presented hybrid methodology.

In [9], they used three models: Adaptive Resonance Theory Based Neural Network (ART), Self-Organizing Map Based Neural Network (SOM), and Back Propagation Neural Network (BPN) in order to classify types of breast cancer. Meanwhile, the BPN method performs better than the other two methods. BPN is one of the types of neural networks that works based on trial-and-error learning and tries to match the given inputs to the outputs by minimizing the value of an error function. Learning in this category of networks is supervised. ART is one of the types of unsupervised networks and is designed to allow the user to control the degree of similarity of the patterns placed in a cluster by adjusting the parameters of vigilance. SOM learning is also unsupervised and uses a competitive learning method for training. Its processing units are regularized in a competitive learning process for input patterns.

In [10], authors designed an expert system for breast cancer detection based on Association Rules (AR) and Neural Networks. Association rules have been used to reduce the dimensions of breast cancer database and Neural Networks for classification. In [11], different types of breast cancer have been classified by applying three methods of 4.5C tree, Multi-Layer Perceptron (MLP), and Naïve Bayes on specific markers of breast tumors. In order to classify the types of Leukemia based on the analysis of gene expression data in [12] CBR learning method was used. In this method, previous examples are remembered, and past experiences are used when facing a new example. Diagnosis of Lymph Nodes Metastases (LNM) before surgery is challenging and complicated. [13], the artificial neural network method was used to diagnose this issue in patients with gastric cancer. In [14], the SVM method was used to classify breast cancer. Since the distinction between benign and malignant tumors is challenging and time-consuming, it seems necessary to use an automated method to detect these cases. In [15], a proposed method can detect and classify cancer from gene expression data using unsupervised deep learning methods. The main advantage of this method is the ability to use data from different types of cancer to automatically form features that help improve the identification and diagnosis of a specific type. In the proposed method, PCA is used to solve the problem of large dimensions of the raw feature space. The authors stated that applying this method to cancer data and comparing it with basic algorithms increased the accuracy of the cancer classification and provided a scalable and general approach to dealing with gene expression data.

In [16], a Sparse Stacked Autoencoder (SSAE) framework is used for automatic nucleus detection in cancer histopathology. The SSAE model can capture the high-level representational of point intensity without supervision. These high-level features have presented a classification that can effectively detect multiple nuclei from a large group of histopathological images. In this paper, the deep structures of SSAE are compared with other AutoEncoders in displaying the high-level features of the intensity of points. In [17], a layered feature selection method is presented along with SSAE. This paper used the Deep Learning method to classify tumors using gene expression data. With the help of SSAEs, high-level features have been extracted from the data. In each layer, a heuristic has been used to obtain relevant features to reduce the fine-tuning of calculations. Ultimately, the classifiers have used the data obtained in the features to perform tumor diagnosis. This paper has tested its presented method on 36 datasets. The datasets are from the GEMLeR source, which includes the gene expression data of 1545 samples from the cells of 9 types of tumors. In the evaluations, the presented method has better results in accuracy and ROC tests on GEMLeR.

In [18], the ability to use deep learning algorithms to detect lung cancer using LIDC images database has been investigated. Each image is divided into parts using the radiologist's marks. After reducing the sampling rate and rotating the images, 174,000 samples with a length and width of 52 pixels were obtained. CNN, DBN, and SDAE have been compared with other methods on a format of 28 image features and a support vector machine. The highest accuracy belongs to the DBN method (0.8119), and the accuracy of the CNN and SDAE methods is equal to 0.7976 and 0.7929, respectively. A vital point in this paper is the size of the images for lung cancer diagnosis, which can cause data loss if the sampling rate decreases. In [19], in 2016, a framework for early detection of prostate cancer from DW-MRI images was presented. The prostate is divided into parts in the first step based on a model trained by a Stochastic Speed function developed from NMF. NMF features are calculated using MRI intensity information, a probabilistic model, and interactions between prostate voxels. In the second step, ADC is obtained for different parts, its values are normalized, and CDF values are created for these prostate cells. In the last step, an AutoEncoder trained by an NCAE algorithm is used to classify prostate tumors as benign or malignant according to the CDFs extracted from the previous step.

In 2015, Deep Learning approaches were used to identify pathology in chest radiograph data [20]. Since large training sets are usually unavailable in the medical domain, the possibility of using deep learning methods based on non-medical training has also been investigated. The proposed algorithm has been tested on a set of 93 images, features extracted from CNN, and a set of features from non-medical domain obtained the best result. The proposed method has shown that deep learning with high-level non-medical image databases can be acceptable for general medical image recognition tasks. [21] introduces an ensemble hierarchical model for combining multiple classifiers. The proposed model consists of two steps: first, a Decision Tree and Logistic Regression models are trained independently, and then their outputs are fed into a Neural Network in the next level. The Neural Network is trained to combine the outputs of the previous classifiers, aiming to improve overall accuracy. The results showed that their proposed model achieved a classification accuracy of over 83%, which outperformed other existing methods in the literature.

In [22], a distributed Deep Learning framework is used to analyze and diagnose diseases. This analysis and diagnosis are based on the questions and answers of doctors and patients. Decision support systems are created using this information. Further, for diagnosis, distributed deep learning method is used to identify features and signatures.

## 3. Proposed Architecture

In this paper, the problem of diagnosis of Leukemia is a classification problem. In this kind of problem, feature extraction has particular importance because it increases data classification accuracy and makes it possible to classify data with large dimensions.
AutoEncoders are multilayer perceptron neural networks whose structure is not optimized for a specific form of data and can accept various types of data regardless of the local, temporal, and serial dependence between their features. AutoEncoders also provides the possibility of feature extraction process in unsupervised learning framework due to the existence of a structure including encoder and decoder. This structure helps the network to be trained in two phases, one phase for feature extraction and the second phase with the feature extractor classifier will be trained again with supervision, increasing the classification's accuracy.

To diagnose Leukemia from the data with the help of AutoEncoders, A structure including three steps of preprocessing, feature extraction, and classification is proposed. Accordingly, apart from machine learning methods used for comparison, this section mentions six experiments conducted to improve and increase classification accuracy. The proposed architecture is shown in Fig. 1, and the input data size, activation functions, optimization algorithms, and the number of layers of AutoEncoder.

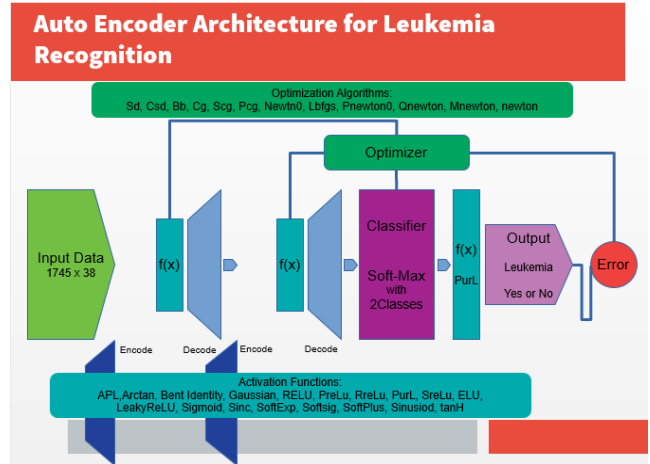

Fig. 1. Proposed Architecture.

### 3.1.  *Preprocessing Data*

Before performing any analysis on the data, it should be normalized, especially when the data is multidimensional. Using normalized data may have an inappropriate effect on the analysis results. Data normalization helps that their importance is independent of their measurement unit. The data were given a normal distribution between zero and one in the preprocessing unit before being given to AutoEncoder. In this test, the data distribution is normalized with the help of z-score standardization method.

### 3.2.  *Feature Extraction*

The model trained in the first attempt had an input size of 38 features, and all the layers of AutoEncoder were given 100 epochs, which was unsupervised training. The whole model is also done with the classifier in 100 epochs for Fine Tune.

The AutoEncoder used in the first experiment has two groups of layers separated from the classifier. The AutoEncoder used in the first layer has 35 neurons, and in the second layer, it has 15 neurons (in the first layer group, the dimensions of the input data is 35 and then returns to the original space. In the second layer group, the input size is 35 will be taken from the hidden layer of the first group and then converted to 15).
The AutoEncoder used in the second experiment and all its attempts had two layers. It has 100 neurons in the first layer and 50 in the second.

The AutoEncoder used in the third experiment has 100 neurons in the first layer, 50 neurons in the second layer, and 25 neurons in the third layer.

The AutoEncoder has four layers in all attempts of the fourth experiment. The AutoEncoder used in the first layer has 100 neurons, the second layer has 50 neurons, the third layer has 25 neurons, and the fourth layer has 12 neurons.

The AutoEncoder used in all attempts of the fifth experiment has 50 neurons in four layers.

In the second test, the L2 weight regularization parameter for each of the network's three layers equals 0.0001. The value of this parameter was the same for all attempts made in the first experiment.

The sparsity proportion in the second experiment in the autoencoder networks is equal, and its value is 0.05. Softmax Regularization in the first test parameter equals 0.0001 for the classifier layer. Stack Regularization is equal to 0.0001 in the first experiment.

In each layer of Autoencoder, activation functions are used to activate each neuron, and parameters of the activation functions used in the structure of Autoencoder in the first experiment are non-adaptive. The activation functions used in the first test are Adaptive Piecewise Liner, arc tangent, Bent Identity, Gaussian, Rectified Liner Unit, Parametric Rectified Liner Unit, Randomized leaky Rectified linear unit, S-shaped Rectified Linear Activation Units, Exponential Liner, Leaky Rectified Linear Unit, sigmoid, sinc, Soft Exponential, Soft sign, Soft Plus, Sinusoid, and Hyperbolic Tangent.

The following optimization methods are used in the model used to train AutoEncoder and minimize the model error. These methods are Steepest Descent, Cyclic Steepest Descent, Barzilai and Borwein Gradient, Nonlinear Conjugate Gradient, Scaled Nonlinear Conjugate Gradient, Preconditioned Nonlinear Conjugate Gradient, Hessian-Free Newton, Newton with Limited-Memory BFGS Updating Quasi, Preconditioned Hessian-Free Newton, Quasi-Newton Hessian approximation, and Newton's Method with Hessian Calculation every iteration.

### 3.3. *Classifier*

(1) Softmax classifier was used to classify the data in the last layer. Also, the applied softmax is trained on 100 epochs and regularized at a rate of 0.001.

## 4. Simulation of Proposed Method

### 1.1. *Dataset*

The first issue in conducting this paper is to collect information on healthy people and patients with Leukemia. Therefore, to obtain the required data, Hamadan Ibn Sina Hospital was referred, and with the cooperation of this hospital, 1745 data samples were prepared and collected. This data sample is related to both genders, of which 55% are related to males, and 45% are related to females. The collected data contains 38 features: Age, gender, having an infection, lymphadenitis in the groin or Inguen (IN.LY), Fever and night sweats (FATH), muscle cramps, swelling of the liver or spleen (LSV), having Swoon, Loose skin, Hemorrhage, Pain bone, Anorexia (W.Irelish), Weight Loss (Depreciatory.bu), Nausea or puke, Cough or asthma, Numbness of the foot (IN.foot), leg swelling (WSF), White blood cell (WBC), Hematocrit, Hemoglobin count, The number of platelets, Sodium count, Lactate dehydrogenase enzyme (LDH), Uric acid, Cratinin, erythrocyte sedimentation rate (ESR), Prothrombin time (PT), PT activated (PTa), patrial thromboplastin time (PTT), Alkaline phosphatase (ALP), Bilirubin total (BillirobinT), Bilirubin direct (BillirobinD), glutamic-oxaloacetic transaminase (SGOT), Serum glutamic pyruvic transaminase (SGPT), mean corpuscular volume (MCV), Mean Corpuscular Hemoglobin (MCH), Mean corpuscular hemoglobin concentration (MCHC), Red blood cell distribution width (RDW), having cancer or not and type of cancer.

**1.2. *Implementation and Evaluation***

The evaluation of the model's efficiency is divided into two parts. In the first phase, the layers of the Autoencoder are trained unsupervised, and the Root Mean Square Error (RMSE) is used as the error calculation criteria. In the second phase, the network is trained supervised, and a softmax classifier is placed at the end. The K-Fold cross-validation method and RMSE are used to evaluate the model. In the first test, the data is divided into five folds.

Based on the results obtained in the first experiment, it was found that the scg algorithm was the best optimization algorithm for training the Autoencoder, and the arctan function was the best activation function according to the optimization algorithm. The second objective was to find the best network structure.

To compare the proposed model with other models, the evaluation criteria used were accuracy (ACC), sensitivity, precision, specificity, Matthew's correlation coefficient (MCC), F1-score, and Receiver Operating Characteristic (ROC).

Table 1. Top 10 implementations with higher Accuracy.

| Layers Neuron Size | Train Algorithm | Activation Function | Evaluation Criteria | | | | | | |
|---|---|---|---|---|---|---|---|---|---|
| | | | *Accuracy* | *RMSE* | *Sensitivity* | *Specificity* | *Precision* | *MCC* | *F1-score* |
| **[30, 15]** | **scg** | **ArcTan** | **0.8297** | **0.1702** | **0.8546** | **0.8546** | **0.8546** | **0.8546** | **0.8546** |
| [30, 15] | pnewton0 | BentIdentity | 0.8286 | 0.1713 | 0.8500 | 0.7986 | 0.8559 | 0.6489 | 0.8525 |
| [30, 15] | newton0 | Softsig | 0.8263 | 0.1736 | 0.8401 | 0.8068 | 0.8598 | 0.6457 | 0.8493 |
| [30, 15] | pcg | Softsig | 0.8229 | 0.1770 | 0.8431 | 0.7944 | 0.8525 | 0.6367 | 0.8476 |
| [30, 15] | lbfgs | Softsig | 0.8217 | 0.1782 | 0.8441 | 0.7903 | 0.8507 | 0.6345 | 0.8469 |
| [30, 15] | cg | TanH | 0.8206 | 0.1793 | 0.8450 | 0.7862 | 0.8481 | 0.6314 | 0.8463 |
| [30, 15] | bb | TanH | 0.8200 | 0.1799 | 0.8627 | 0.7600 | 0.8355 | 0.6282 | 0.8485 |
| [30, 15] | lbfgs | ArcTan | 0.8194 | 0.1805 | 0.8470 | 0.7806 | 0.8470 | 0.6310 | 0.8456 |
| [30, 15] | pcg | ArcTan | 0.8166 | 0.1833 | 0.8215 | 0.8096 | 0.8604 | 0.6292 | 0.8456 |
| [30, 15] | newton0 | Sigmoid | 0.8166 | 0.1833 | 0.8333 | 0.7931 | 0.8504 | 0.6257 | 0.8411 |

According to Table 1, the best optimization algorithm for training is the scg, and the best activation function is ArcTan is selected and used in the following experiment. Then by changing the number of layers of Autoencoder in the following experiments, we want to find the best network structure.

Table 2. Results of AutoEncoder in Each Experiments.

| Layers Neuron Size | Evaluation Criteria | | | | | | |
|---|---|---|---|---|---|---|---|
| | *Accuracy* | *RMSE* | *Sensitivity* | *Specificity* | *Precision* | *MCC* | *F1-score* |
| [100, 50] | 0.8200 | 0.1799 | 0.8509 | 0.7765 | 0.8439 | 0.6320 | 0.8461 |
| [100, 50, 25] | 0.8240 | 0.1759 | 0.8441 | 0.7958 | 0.8542 | 0.6409 | 0.8482 |
| [100, 50, 25, 12] | 0.8246 | 0.1753 | 0.8539 | 0.7834 | 0.8486 | 0.6409 | 0.8502 |
| [50, 50, 50, 50] | 0.8246 | 0.1753 | 0.8480 | 0.7917 | 0.8520 | 0.6406 | 0.8494 |
| **[1024 512 250 100]** | **0.8275** | **0.1724** | **0.8696** | **0.7682** | **0.8414** | **0.6442** | **0.8547** |

After finding our architecture for AutoEncoder, we compare it to the classical machine learning method used in this area. According to the experiments conducted, AutoEncoder has a higher average accuracy among the classical methods, and the PCA-SVM method is in the next rank among the classical methods. These results are shown in Table 3 and Fig 2.

Table 3. Results of Comparing Proposed Method to Others.

| Model | Evaluation Criteria | |
|---|---|---|
| | *Accuracy* | *F1-score* |
| PCA – 1NN | 70.87% | 73.46% |
| PCA – 3NN | 69.64% | 71.82% |
| PCA – 5NN | 67.46% | 69.88% |
| PCA – SVM | 71.04% | 73.96% |
| PCA - RBF | 66.32% | 68.03% |
| PCA - SoftMax | 70.16% | 72.78% |
| **Proposed Model** | **82.97%** | **85.68%** |

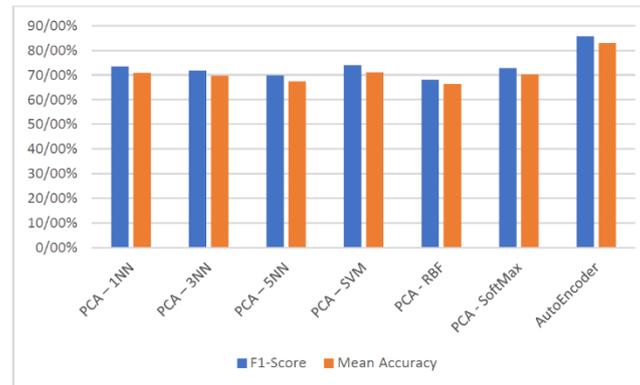

Fig. 2. The results of machine learning and AutoEncoder model.

## 5. Conclusion

In Leukemia, cancer cells multiply rapidly, and the blood can no longer perform its functions. Early recognition of Leukemia helps patients to start treatment without interruption after diagnosis. In past research, machine learning methods have been used to diagnose cancer, each of which had average accuracy. With the introduction of AutoEncoders and increasing the accuracy of these networks, an attempt has been made to use them for Leukemia diagnosis. In this paper, the samples of the patients of Sina Hospital in Hamadan were used. Data preprocessing and normalization is the first step in using the dataset in this study. In this paper, the first question is to find the best optimization algorithm for training the weights of the AutoEncoder network, which was the SCG algorithm, and also the best activation function according to the optimization algorithm, which was ArcTan. The second step was finding the best structure of the network. Different tests by changing the layers of AutoEncoder resulted in two hidden layers: the number of neurons in the first was 30, and the number of neurons in the second was 15. The correctness of this network with the mean squared error of 0.17 has reached the average accuracy value of 82.97%, and its f1-Score is equal to 85.68%.